\documentclass[11pt,a4paper]{article}
\usepackage[hyperref]{emnlp2018}

\usepackage{times}
\usepackage{latexsym}

\usepackage{url}

\usepackage{array}
\usepackage{amsmath}
\usepackage{graphicx}
\usepackage{bbm}
\usepackage{natbib}

\usepackage{multirow}
\usepackage{multicol}

\usepackage{booktabs}

\usepackage{appendix}

\aclfinalcopy 

\newcommand{\ba}{\mathbf{a}}
\newcommand{\bm}{\mathbf{m}}

\newcommand{\be}{\mathbf{e}}

\newcommand{\bh}{\mathbf{h}}

\newcommand{\bA}{\mathbf{A}}

\newcommand{\br}{\mathbf{r}}

\newcommand{\mE}{\mathcal{E}}
\newcommand{\mG}{\mathcal{G}}
\newcommand{\mH}{\mathcal{H}}

\newcommand{\mR}{\mathcal{R}}

\newcommand{\mS}{\mathcal{S}}
\newcommand{\mA}{\mathcal{A}}

\newcommand{\one}{\mathbbm{1}}
\newcommand{\zero}{\mathbf{0}}
\newcommand{\Real}{\mathbbm{R}}
\newcommand{\expect}[2]{\mathbbm{E}_{#1}[#2]}

\newcommand{\LSTM}{\text{LSTM}}
\newcommand{\softmax}[1]{\sigma(#1)}
\newcommand{\relu}[1]{\text{ReLU}(#1)}

\newcommand{\sref}[1]{\S\ref{#1}}

\newcommand{\grad}[2]{\nabla_#1#2}
\newcommand{\policy}{\pi_{\theta}(a_t|s_t)}

\newcommand{\tworow}[1]{\multirow{2}{*}{#1}}
\newcommand{\twocol}[1]{\multicolumn{2}{c}{#1}}
\newcommand{\threecol}[1]{\multicolumn{3}{c}{#1}}
\newcommand{\fourcol}[1]{\multicolumn{4}{c|}{#1}}
\newcommand{\fourcolnb}[1]{\multicolumn{4}{c}{#1}}

\newcommand{\highest}[1]{\textbf{#1}}
\newcommand{\rela}[2]{#1 (#2\%)}

\newcommand{\edge}[1]{$-$#1$\rightarrow$}
\newcommand{\redge}[1]{$\leftarrow$#1$-$}

\newcommand\twohopr[3]{\textit{#1 $\wedge$ #2 $\Rightarrow$ #3}}

\title{Multi-Hop Knowledge Graph Reasoning with Reward Shaping}

\author{Xi Victoria Lin \qquad
  Richard Socher \qquad
  Caiming Xiong \\
  Salesforce Research \\
  {\tt \{xilin,rsocher,cxiong\}@salesforce.com}
}
\date{}

\begin{document}
\maketitle
\begin{abstract}
Multi-hop reasoning is an effective approach for query answering (QA) over incomplete knowledge graphs (KGs). 
The problem can be formulated in a reinforcement learning (RL) setup, where a policy-based agent sequentially extends its inference path until it reaches a target.
However, in an incomplete KG environment, the agent receives low-quality rewards corrupted by false negatives in the training data, which harms generalization at test time. Furthermore, since no golden action sequence is used for training, the agent can be misled by spurious search trajectories that incidentally lead to the correct answer.
We propose two modeling advances to address both issues: (1) we reduce the impact of false negative supervision by adopting a pretrained one-hop embedding model to estimate the reward of unobserved facts; (2) we counter the sensitivity to spurious paths of on-policy RL by forcing the agent to explore a diverse set of paths using randomly generated edge masks.
Our approach significantly improves over existing path-based KGQA models on several benchmark datasets and is comparable or better than embedding-based models.


\end{abstract}

\section{Introduction}
\label{sec:intro}

Large-scale knowledge graphs (KGs) support a variety of downstream NLP applications such as semantic search~\cite{WebQuestions} and dialogue generation~\cite{CoCoA}. 
Whether curated automatically or manually, practical KGs often fail to include many relevant facts. 
A popular approach for modeling incomplete KGs is knowledge graph embeddings, which map both entities and relations in the KG to a vector space and learn a truth value function for any potential KG triple parameterized by the entity and relation vectors~\cite{DistMult,ConvE}.

\begin{figure}[t]
	\centering
	\includegraphics[width=\linewidth]{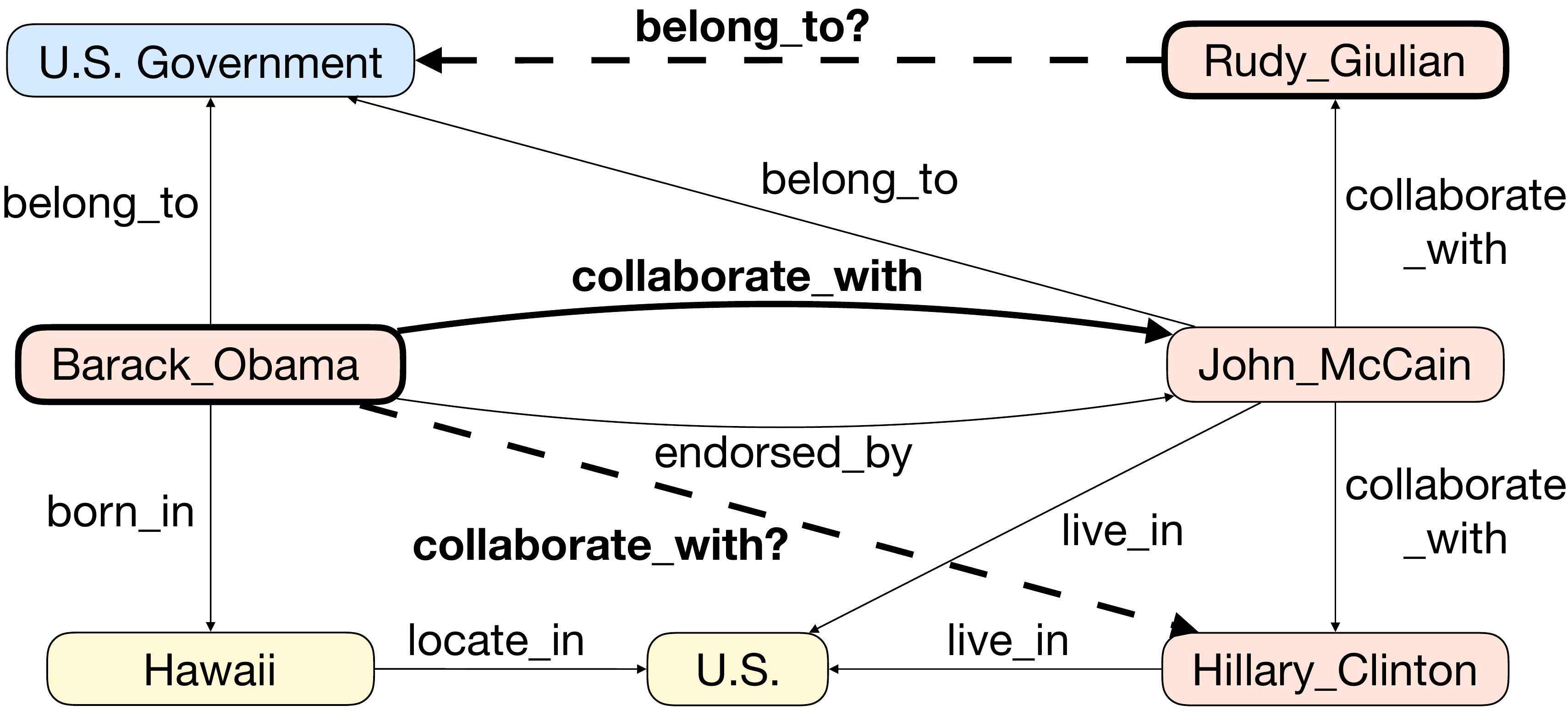}
	\caption{Example of an incomplete knowledge graph which 
    contains missing links (dashed lines) that can possibly be inferred from existing facts (solid lines).
    }
	\label{fig:kg}
\end{figure}

Embedding based approaches ignore the symbolic compositionality of KG relations, which limit their application in more complex reasoning tasks. An alternative solution for KG reasoning is to infer missing facts by synthesizing information from multi-hop paths, e.g. \twohopr{bornIn(Obama, Hawaii)}{locatedIn(Hawaii, US)}{bornIn(Obama, US)}, as shown in Figure~\ref{fig:kg}.
Path-based reasoning offers logical insights of the underlying KG and are more directly interpretable. 
Early work treats it as a link prediction problem and perform maximum-likelihood classification over either discrete path features~\cite{PRA,DBLP:conf/emnlp/LaoSPC12,DBLP:conf/emnlp/GardnerTKM13} or their hidden representations in a vector space~\cite{TraverseKG,DBLP:conf/acl/ToutanovaLYPQ16,ChainofReason}. 

More recent work formulates multi-hop reasoning as a sequential decision problem, and leverages reinforcement learning (RL) to perform effective path search~\cite{DeepPath,minerva,ReinforceWalk,VariationalKB}. 
In particular, MINERVA~\cite{minerva} uses 
the REINFORCE algorithm~\cite{REINFORCE} to train an end-to-end model for multi-hop KG query answering: given a query relation and a source entity, the trained agent searches over the KG starting from the source and arrives at the candidate answers without access to any pre-computed paths. 

\begin{figure}[t]
	\centering
	\includegraphics[width=.95\linewidth]{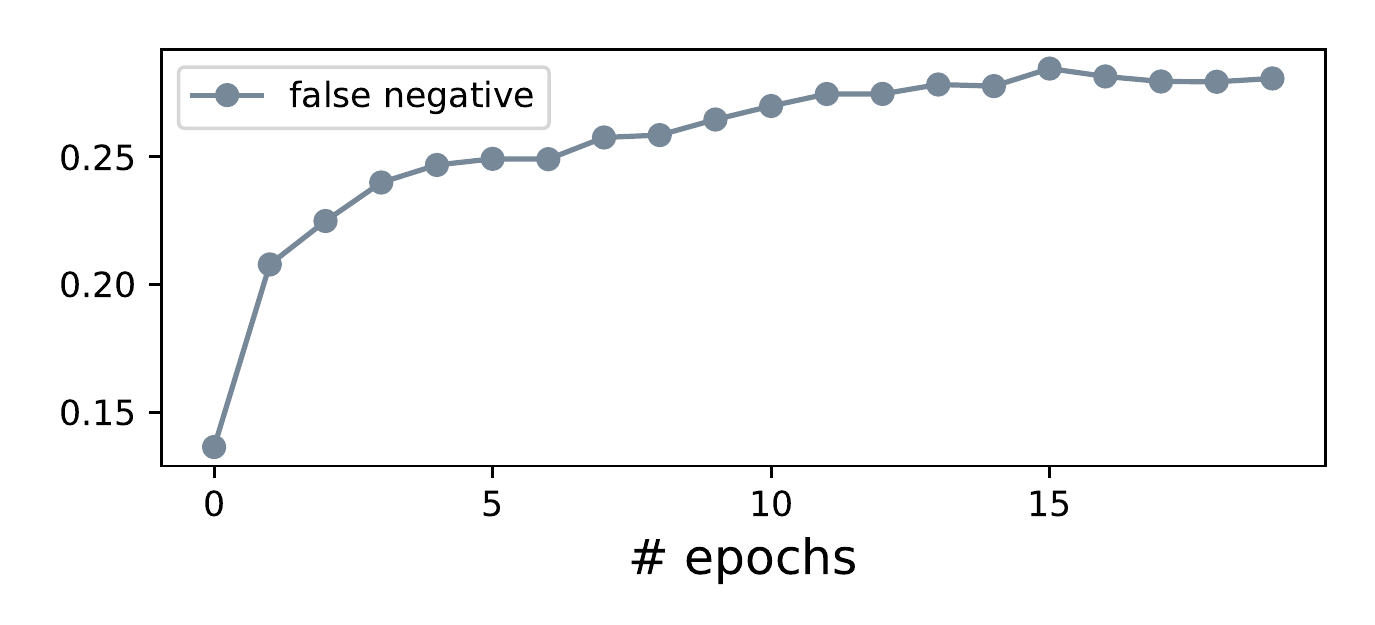}
	\caption{ 
    	Percentage of false negatives hit (where the model predicted an answer that exists in the full KG but cannot be identified by the training subset) in the first 20 epochs of walk-based QA training on the UMLS knowledge graph~\cite{DBLP:conf/icml/KokD07}.}
	\label{fig:fn}
\end{figure}

We refer to the RL formulation adopted by MINERVA as ``learning to walk towards the answer'' or ``walk-based query-answering (QA)''. 
Walk-based QA eliminates the need to pre-compute path features, yet 
this setup poses several challenges for training. 
First, because practical KGs are intrinsically incomplete, the agent may arrive at a correct answer whose link to the source entity is missing from the training graph without receiving any reward (false negative targets, Figure~\ref{fig:fn}).
Second, since no ground truth path is available for training, the agent may traverse spurious paths that lead to a correct answer only incidentally (false positive paths). 
Because REINFORCE~\cite{REINFORCE} is an on-policy~\cite{DBLP:books/lib/SuttonB98} RL algorithm which encourages past actions with high reward, it can bias the policy toward spurious paths found early in training ~\cite{RLMML}.

We propose two modeling advances for RL approaches in the walk-based QA framework to address the previously mentioned problems. First, instead of using a binary reward based on whether the agent has reached a correct answer or not, we adopt pre-trained state-of-the-art embedding-based models~\cite{ConvE,ComplEx} to estimate a {\bf\emph{soft reward}} for target entities whose correctness cannot be determined. 
As embedding-based models capture link semantics well, unobserved but correct answers would receive a higher reward score compared to a true negative entity using a well-trained model.
Second, we perform {\bf\emph{action dropout}} which randomly blocks some outgoing edges of the agent at each training step so as to enforce effective exploration of a diverse set of paths and dilute the negative impact of the spurious ones.  
Empirically, our overall model significantly improves over state-of-the-art multi-hop reasoning approaches on four out of five benchmark KG datasets (UMLS, Kinship, FB15k-237, WN18RR). 
It is also the first path-based model 
that achieves consistently comparable or better performance than embedding-based models.
In addition, we perform a thorough ablation study and result analysis, demonstrating the effect of each modeling innovation. 




\section{Approach}
\label{sec:bg}
In this section, we first review the walk-based QA framework (\S\ref{subsec:minerva}) and the on-policy reinforcement learning approach proposed by~\citet{minerva} (\S\ref{subsec:pg},\S\ref{subsec:opt}).
Then we describe our proposed solutions to the false negative reward and spurious path problems: knowledge-based reward shaping (\S\ref{subsec:rs}) and action dropout (\S\ref{subsec:ad}).

\subsection{Formal Problem Definition}
\label{subsec:notation}
We formally represent a knowledge graph as $\mG = (\mE, \mR)$, where $\mE$ is the set of entities and $\mR$ is the set of relations. Each directed link in the knowledge graph $l = (e_s, r, e_o)\in\mG$ represents a fact (also called a triple). 

Given a query $(e_s, r_q, ?)$, where $e_s$ is the source entity and $r_q$ is the relation of interest, the goal is to perform an efficient search over $\mG$ and collect the set of possible answers $E_o=\{e_o\}$ s.t. $(e_s, r_q, e_o)\notin\mG$ due to incompleteness.

\subsection{Reinforcement Learning Formulation}
\label{subsec:minerva}

The search can be viewed as a Markov Decision Process (MDP)~\cite{DBLP:books/lib/SuttonB98}: starting from $e_s$, the agent sequentially selects an outgoing edge $l$ and traverses to a new entity until it arrives at a target. Specifically, the MDP consists of the following components~\cite{minerva}.

\paragraph{States} Each state $s_t=(e_t, (e_s, r_q))\in\mS$ is a tuple where $e_t$ is the entity visited at step $t$ and $(e_s, r_q)$ are the source entity and query relation. 
$e_t$ can be viewed as state-dependent information while $(e_s, r_q)$ are the global context shared by all states.

\paragraph{Actions} The set of possible actions $A_t\in\mA$ of at step $t$ consists of the outgoing edges of $e_t$ in $\mG$. Concretely, $A_t=\{(r', e') | (e_t, r', e')\in\mG\}$. 
To give the agent the option of terminating a search, a self-loop edge is added to every $A_t$. 
Because search is unrolled for a fixed number of steps $T$, the self-loop acts similarly to a ``stop'' action.

\paragraph{Transition} A transition function $\delta: \mS\times\mA\rightarrow\mS$ is defined by $\delta(s_t, A_t)=\delta(e_t, (e_s, r_q), A_t)$. 
For walk-based QA, the transition is entirely determined by $\mG$.

\paragraph{Rewards} In the default formulation, the agent receives a terminal reward of 1 if it arrives at a correct target entity at the end of search and 0 otherwise. 
\begin{equation}
\label{eq:breward}
	R_b(s_T) = \one\{(e_s, r_q, e_T)\in\mG\}.
\end{equation}

\subsection{Policy Network}
\label{subsec:pg}
The search policy is parameterized using state information and global context, plus the search history~\cite{minerva}.

Specifically, every entity and relation in $\mG$ is assigned a dense vector embedding $\be\in\Real^d$ and $\br\in\Real^d$. The action $a_t=(r_{t+1}, e_{t+1})\in A_t$ is represented as the concatenation of the relation embedding and the end node embedding $\ba_t = [\br; \be_t']$.

The search history $h_t= (e_s, r_1, e_1, \dots, r_t, e_t)\in\mH$ consists of the sequence of observations and actions taken up to step $t$, and can be encoded using an LSTM:
\begin{align}
\label{eq:history}
	\bh_0 &= \LSTM(\zero, [\br_0; \be_s]) \\
    \bh_t &= \LSTM(\bh_{t-1}, \ba_{t-1}),\ t > 0,
\end{align}
where $r_0$ is a special start relation introduced to form a start action with $e_s$. 

The action space is encoded by stacking the embeddings of all actions in $A_t$: $\bA_t\in\Real^{|A_t|\times 2d}$. And the policy network $\pi$ is defined as:
\begin{align}
\label{eq:mlp}
	\pi_{\theta}(a_t|s_t) = \softmax{\bA_t \times W_2~\relu{W_1[\be_t;\bh_t;\br_q]}},
\end{align}
where $\sigma$ is the softmax operator. 

\subsection{Optimization}
\label{subsec:opt}
The policy network is trained by maximizing the expected reward over all queries in $\mG$:
\begin{equation}
	\label{eq:rlobjective}
    J(\theta) = \expect{(e_s,r,e_o)\in\mG}{\expect{a_1,\dots,a_T\sim \pi_{\theta}}{R(s_T|e_s, r)}}. 
\end{equation}
The optimization is done using the REINFORCE~\cite{REINFORCE} algorithm, which iterates through all $(e_s, r, e_o)$ triples in $\mG$\footnote{This training strategy treats a query with $n>1$ answers as $n$ single-answer queries. In particular, given a query $(e_s, r_q, ?)$ with multiple answers $\{e_{t_1},\hdots e_{t_n}\}$, when training w.r.t. the example $(e_s, r_q, e_{t_i})$, MINERVA removes all $\{e_{t_j}|j\neq i\}$ observed in the training data from the possible set of target entities in the last search step so as to force the agent to walk towards $e_{t_i}$. We adopt the same technique in our training.} and updates $\theta$ with the following stochastic gradient:
\begin{equation}
\label{eq:gupdate}
   \grad{\theta}{J(\theta)} \approx \grad{\theta}{\sum_tR(s_T|e_s, r)\log{\pi_{\theta}(a_t|s_t)}}. 
\end{equation}

\subsection{Knowledge-Based Reward Shaping}
\label{subsec:rs}
According to equation~\ref{eq:breward}, the agent receives a binary reward based on solely the observed answers in $\mG$. However, $\mG$ is intrinsically incomplete and this approach rewards the false negative search results identically to true negatives.
To alleviate this problem, we use existing KG embedding models designed for the purpose of KG completion~\cite{ComplEx,ConvE} to estimate a soft reward for target entities whose correctness is unknown.

Formally, the embedding models map $\mE$ and $\mR$ to a vector space, and estimate the likelihood of each fact $l = (e_s, r, e_t)\in\mG$ using a composition function of the entity and relation embeddings $f(e_s, r, e_t)$. 
$f$ is trained by maximizing the likelihood of all facts in $\mG$. 
We propose the following reward shaping strategy~\cite{DBLP:conf/icml/NgHR99}:
\begin{equation}
	\label{eq:sreward}
    R(s_T) = R_b(s_T) + (1 - R_b(s_T))f(e_s, r_q, e_T).
\end{equation}
Namely, if the destination $e_T$ is a correct answer according to $\mG$, the agent receives reward 1. Otherwise the agent receives a fact score estimated by $f(e_s, r_q, e_T)$, which is pre-trained. 
Here we keep $f$ in its general form and it can be replaced by any state-of-the-art model~\cite{ComplEx,ConvE} or ensemble thereof. 


\begin{figure*}[t]
	\centering
	\includegraphics[width=.94\linewidth]{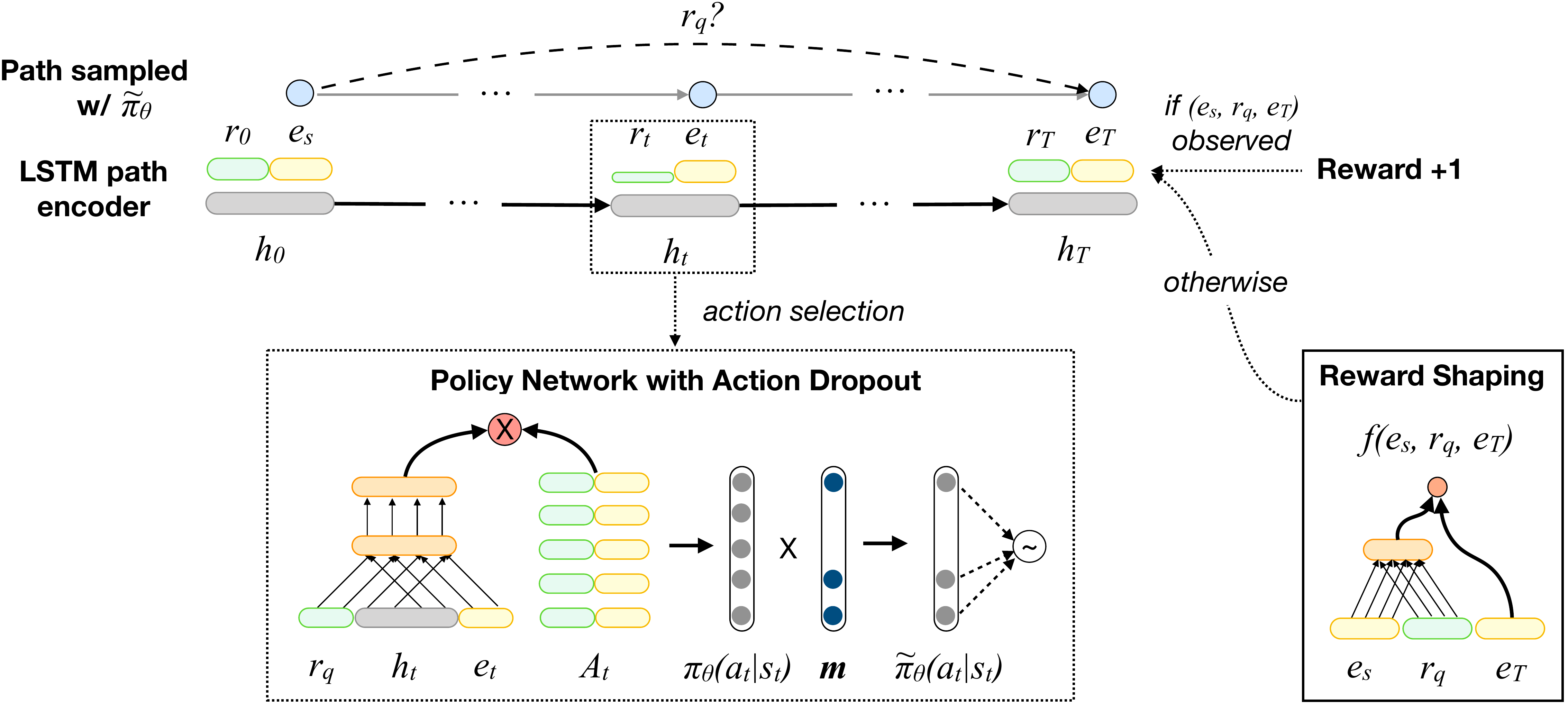}
	\caption{Overall training approach. At each time step $t$, the agent samples an outgoing link according to $\tilde{\pi}_{\theta}(a_t|s_t)$, which is the stochastic REINFORCE policy $\pi_{\theta}(a_t|s_t)$ perturbed by a random binary mask $\bm$. The agent receives reward 1 if stopped at an observed answer of the query $(e_s, r_q, ?)$; otherwise, it receives reward $f(e_s,r_q,e_T)$ estimated by the reward shaping (RS) network. The RS network is pre-trained and doesn't receive gradient updates.
    }
	\label{fig:approach}
\end{figure*}

\subsection{Action Dropout}
\label{subsec:ad}
The REINFORCE training algorithm performs on-policy sampling according to $\pi_{\theta}(a_t|s_t)$, and updates $\theta$ stochastically using equation~\ref{eq:gupdate}. Because the agent does not have access to any oracle path, it is possible for it to arrive at a correct answer $e_o$ via a path which is irrelevant to the query relation. As shown in Figure~\ref{fig:kg}, the path \emph{Obama \edge{endorsedBy} McCain \edge{liveIn} U.S. \redge{locatedIn} Hawaii} does not infer the fact $bornIn(Obama, Hawaii)$. 

Discriminating paths of different qualities is non-trivial, and existing RL approaches for walk-based KGQA largely rely on the terminal reward to bias the search. Since there are usually more spurious paths than correct ones, spurious paths are often found first, and following exploration can be increasingly biased towards them (Equation~\ref{eq:gupdate}).
Entities with larger fan-in (in-degree) and fan-out (out-degree) often exacerbate this problem.

\citet{RLMML} identified a similar issue in RL-based semantic parsing with weak supervision, where programs that do not semantically match the user utterance frequently pass the tests. 
To solve this problem, \citet{RLMML} proposed randomized beam search combined with a meritocratic update rule to ensure all trajectories that obtain rewards are up-weighted roughly equally.

Here we propose the action dropout technique which achieves similar effect as randomized search and is simpler to
implement over graphs. Action dropout randomly masks some outgoing edges for the agent in the sampling step of REINFORCE. The agent then performs sampling\footnote{We only modify the sampling distribution and still use $\policy$ to compute the gradient update in equation~\ref{eq:gupdate}.} according to the adjusted action distribution
\begin{align}
\label{eq:ad}
& \tilde{\pi}_{\theta}(a_t|s_t) \propto (\policy \cdot \bm + \epsilon) \\
& m_i \sim Bernoulli(1 - \alpha), i = 1,\dots|A_t|,
\end{align}
where each entry of $\bm\in\{0,1\}^{|A_t|}$ is a binary variable sampled from the Bernoulli distribution with parameter $1-\alpha$.
A small value $\epsilon$ is used to smooth the distribution in case $\bm=\zero$, where $\tilde{\pi}_{\theta}(a_t|s_t)$ becomes uniform.

Our overall approach is illustrated in figure~\ref{fig:approach}.
\section{Related Work}

In this section, we summarize the related work and discuss their connections to our approach.

\subsection{Knowledge Graph Embeddings}
KG embeddings~\cite{TransE,NTN,DistMult,ComplEx,ConvE} are one-hop KG modeling approaches which learn a scoring function $f(e_s, r, e_o)$ to define a fuzzy truth value of a triple in the embedding space. These models can be adapted for query answering by simply return the $e_o$'s with the highest $f(e_s, r, e_o)$ scores. Despite their simplicity, embedding-based models achieved state-of-the-art performance on KGQA~\cite{minerva}. 
However, such models ignore the symbolic compositionality of KG relations, which limits their usage in more complex reasoning tasks. 
The reward shaping (RS) strategy we proposed is a step to combine their capability in modeling triple semantics with the symbolic reasoning capability of the path-based approach.

\subsection{Multi-Hop Reasoning}
Multi-hop reasoning focus on learning symbolic inference rules from relational paths in the KG and has been formulated as sequential decision problems in recent works~\cite{DeepPath,minerva,ReinforceWalk,VariationalKB}.
In particular, DeepPath~\cite{DeepPath} first adopted REINFORCE to search for generic representative paths between pairs of entities. \textsc{Diva}~\cite{VariationalKB} also performs generic path search between entities using RL and its variational objective can be interpreted as model-based reward assignment.
MINERVA~\cite{minerva} first introduced RL to search for answer entities of a particular KG query end-to-end. MINERVA uses entropy regularization to softly encourage the policy to sample diverse paths, and we show that hard action dropout is more effective in this setup.
ReinforceWalk~\cite{ReinforceWalk} further proposed to solve the reward sparsity problem in walk-based QA using off-policy learning. ReinforceWalk scores the search targets with a value function which is updated based on the search history accumulated through epochs. In comparison, we leveraged existing embedding-based models for reward shaping, which makes the training more efficient.   

\subsection{Reinforcement Learning}
Recently, RL has seen a variety of applications in NLP including machine translation~\cite{DBLP:journals/corr/RanzatoCAZ15}, summarization~\cite{DBLP:journals/corr/PaulusXS17}, and semantic parsing~\cite{RLMML}. Compared to the domain of games~\cite{DBLP:journals/corr/MnihKSGAWR13} and many other applications, RL formulations in NLP often have a large action space (e.g., in machine translation, the space of possible actions is the entire vocabulary of a language). This also holds for KGs, as some entities may have thousands of neighbors (e.g. $U.S.$). Since often there is no golden path available for a KG reasoning problem, we cannot use supervised pre-training to give the path search a better start position following the common practice adopted in RL-based natural language generation~\cite{DBLP:journals/corr/RanzatoCAZ15}. On the other hand, the inference paths being studied in a KG are often much shorter (usually containing 2-5 steps) compared to the NL sentences in the sequence generation problems (often containing 20-30 words). 
\section{Experiment Setup}
We evaluate our modeling contributions on five KGs from different domains and exhibiting different graph properties (\S~\ref{subsec:data}). 
We compare with two classes of state-of-the-art KG models: multi-hop neural symbolic approaches and KG embeddings (\sref{subsec:baseline}).
In this section, we describe the datasets and our experiment setup in detail.

\subsection{Dataset}
\label{subsec:data}

We adopt five benchmark KG datasets for query answering: (1) Alyawarra Kinship, (2) Unified Medical Language Systems~\cite{DBLP:conf/icml/KokD07}, (3) FB15k-237~\cite{DBLP:conf/emnlp/ToutanovaCPPCG15}, (4) WN18RR~\cite{ConvE}, and (5) NELL-995~\cite{DeepPath}. The statistics of the datasets are shown in Table~\ref{tb:data}. 

\begin{table}
    \small
    \centering
    \setlength\tabcolsep{4pt}
    \begin{tabular}{lrrrrr}
    	\hline
    	\tworow{\textbf{Dataset}} & \tworow{\textbf{\#Ent}} & \tworow{\textbf{\#Rel}} & \tworow{\textbf{\#Fact}} & \twocol{\textbf{\#degree}} \\
        \cline{5-6}
        & & & & \textbf{mean} & \textbf{median} \\
        \hline 
        Kinship & 104 & 25 & 8,544 & 85.15 & 82 \\
        UMLS & 135 & 46 & 5,216 & 38.63 & 28 \\
        FB15k-237 & 14,505 & 237 & 272,115 & 19.74 & 14 \\
        WN18RR & 40,945 & 11 & 86,835 & 2.19 & 2 \\
        NELL-995 & 75,492 & 200 & 154,213 & 4.07 & 1 \\
        \hline
    \end{tabular}
    \caption{KGs used in the experiments sorted by increasing sparsity level.}
    \label{tb:data}
\end{table}
\begin{table*}[ht]
    \centering
    \hspace*{-0.32cm}
    \setlength\tabcolsep{3.1pt}
    \setlength\extrarowheight{3pt}
    \scalebox{0.85}{
    \begin{tabular}{l|ccc|ccc|ccc|ccc|ccc}
    	\hline
    	\tworow{\textbf{Model}} & \threecol{\textbf{UMLS}} & \threecol{\textbf{Kinship}} & \threecol{\textbf{FB15k-237}} & \threecol{\textbf{WN18RR}} & \threecol{\textbf{NELL-995}} \\
        \cline{2-4}\cline{5-7}\cline{8-10}\cline{11-13}\cline{14-16} 
        & \small{@1} & \small{@10} & \small{MRR} & \small{@1} & \small{@10} & \small{MRR} & \small{@1} & \small{@10} & \small{MRR} & \small{@1} & \small{@10} & \small{MRR} & \small{@1} & \small{@10} & \small{MRR} \\
        \hline
        DistMult~\cite{DistMult} & 82.1 & 96.7 & 86.8 & 48.7 & 90.4 & 61.4 & 32.4 & 60.0 & 41.7 & \highest{43.1} & 52.4 & \highest{46.2} & 55.2 & 78.3 & 64.1 \\
        ComplEx~\cite{ComplEx} & 89.0 & 99.2 & 93.4 & \highest{81.8} & \highest{98.1} & \highest{88.4} & 32.8 & 61.6 & 42.5 & 41.8 & 48.0 & 43.7 & 64.3 & 86.0 & 72.6 \\
        ConvE~\cite{ConvE} & \highest{93.2} & \highest{99.4} & \highest{95.7} & 79.7 & \highest{98.1} & 87.1 & \highest{34.1} & \highest{62.2} & \highest{43.5} & 40.3 & \highest{54.0} & 44.9 & \highest{67.8} & \highest{88.6} & \highest{76.1} \\
        \hline
        NeuralLP~\cite{NeuralLP} & 64.3 & 96.2 & 77.8 & 47.5 & 91.2 & 61.9 & 16.6 & 34.8 & 22.7 & 37.6 & \highest{65.7} & 46.3 & -- & -- & -- \\
        NTP-$\lambda$ (Rockt{\"{a}}schel et. al. 2017) & 84.3 & \highest{100} & 91.2 & 75.9 & 87.8 & 79.3 & -- & -- & -- & -- & -- & -- & -- & -- & -- \\
        MINERVA~\cite{minerva} & 72.8 & 96.8 & 82.5 & 60.5 & 92.4 & 72.0 & 21.7 & 45.6 & 29.3 & 41.3 & 51.3 & 44.8 & \highest{66.3} & 83.1 & 72.5 \\
        Ours(ComplEx) & 88.7 & 98.5 & 92.9 & \highest{81.1} & \highest{98.2} & \highest{87.8} & \highest{32.9} & 54.4 & 39.3 & \highest{43.7} & 54.2 & \highest{47.2} & 65.5 & 83.6 & 72.2 \\
        Ours(ConvE) & \highest{90.2} & 99.2 & \highest{94.0} & 78.9 & \highest{98.2} & 86.5 & 32.7 & \highest{56.4} & \highest{40.7} & 41.8 & 51.7 & 45.0 & 65.6 & \highest{84.4} & \highest{72.7} \\ 
        \bottomrule
    \end{tabular}
    }
    \caption{Query answering performance compared to state-of-the-art embedding based approaches (top part) and multi-hop reasoning approaches (bottom part). The @1, @10 and MRR metrics were multiplied by 100. We highlight the best approach in each category. 
    \label{tb:mainresults}
    }
\end{table*}

\subsection{Baselines and Model Variations} 
\label{subsec:baseline}

We compare with three embedding based models: DistMult~\cite{DistMult}, ComplEx~\cite{ComplEx} and ConvE~\cite{ConvE}. We also compare with three multi-hop neural symbolic models: (a) NTP-$\lambda$, an improved version of Neural Theorem Prover~\cite{NTP}, (b) Neural Logical Programming (NeuralLP)~\cite{NeuralLP} and (c) MINERVA. 
For our own approach, we include two model variations that use ComplEx and ConvE as the reward shaping modules respectively, denoted as Ours(ComplEx) and Ours(ConvE). We quote the results of NeuralLP, NTP-$\lambda$ and MINERVA reported in~\citet{minerva}, and replicated the embedding based systems.\footnote{~\citet{minerva} reported MINERVA results with the entity embedding usage as an extra hyperparameter -- the quoted performance of MINERVA in Table~\ref{tb:mainresults} on UMLS and Kinship were obtained with entity embeddings setting to zero. In contrast, our system always uses trained entity embeddings.}

\subsection{Implementation Details}
\label{subsec:details}

\paragraph{Beam Search Decoding} We perform beam search decoding to obtain a list of unique entity predictions. Because multiple paths may lead to the same target entity, we compute the list of unique entities reached in the final search step and assign each of them the maximum score among all paths that led to it. We then output the top-ranked unique entities. We find this approach to improve over outputting the entities ranked at the beam top directly, as many of them are repetitions.

\paragraph{KG Setup} Following previous work, we treat every KG link as bidirectional and augment the graph with the reversed $(e_o, r^{-1}, e_s)$ links.  We use the same train, dev, and test set splits as~\citet{minerva}. 
We exclude any link from the dev and test set (and its reversed link) from the train set in case there is an overlap.
Following~\citet{minerva}, we cut the maximum number of outgoing edges of an entity by a threshold $\eta$ to prevent GPU memory overflow: for each entity we retain its top-$\eta$ neighbors with the highest PageRank scores~\cite{PageRank}.

\paragraph{Hyperparameters} We set the entity and relation embedding size to 200 for all models.
We use Xavier initialization~\cite{DBLP:journals/jmlr/GlorotB10} for the embeddings and the NN layers. For ConvE, we use the same convolution layer and label smoothing hyperparameters as~\citet{ConvE}. For path-based models, we use a three-layer LSTM as the path encoder and set its hidden dimension to 200. We perform grid search on the reasoning path length ($2, 3$), the node fan-out threshold $\eta$ ($256$-$512$) and the action dropout rate $\alpha$ ($0.1$-$0.9$). Following~\citet{minerva}, we add an entropy regularization term in the objective and tune the weight parameter $\beta$ within 0-0.1. We use Adam optimization~\cite{Adam} and search the learning rate ($0.001$-$0.003$) and mini-batch size ($128$-$512$).\footnote{On some datasets, we found larger batch size to continue improving the performance but had to stop at 512 due to memory constraints.} For all models we apply dropout to the entity and relation embeddings and all feed-forward layers, and search the dropout rates within $0$-$0.5$. We use a decoding beam size of 512 for NELL-995 and 128 for the other datasets. 

\paragraph{Evaluation Protocol} We convert each triple $(e_s, r, e_o)$ in the test set into a query and compute ranking-based evaluation metrics. The models take $e_s, r$ as the input and output a list of candidate answers $E_o=[e^1,\dots,e^L]$ ranked in decreasing order of confidence score. We compute $r_{e_o}$, the rank of $e_o$ among $E_o$, after removing the other correct answers from $E_o$ and use it to compute two types of metrics: (1) Hits@$k$ which is the percentage of examples where $r_{e_o} \leq k$ and (2) mean reciprocal rank (MRR) which is the mean of $1/r_{e_o}$ for all examples in the test set. 
We use the entire test set for evaluation, with the exception of NELL-995, where test triples with unseen entities are removed following~\citet{minerva}.

We will release the Pytorch implementation of all experiments. Please check the authors' web page for updates.
\section{Results}
\label{subsec:results}

\subsection{Model Comparison}
Table~\ref{tb:mainresults} shows the evaluation results of our proposed approach and the baselines. The top part presents embedding based approaches and the bottom part presents multi-hop reasoning approaches.\footnote{We report the model robustness measurements in \S~\ref{ap:model_robust}.}

We find embedding based models perform strongly on several datasets, achieving overall best evaluation metrics on UMLS, Kinship, FB15K-237 and NELL-995 despite their simplicity. While previous path based approaches achieve comparable performance on some of the datasets (WN18RR, NELL-995, and UMLS), the performance gaps to the embedding based models on the other datasets (Kinship and FB15k-237) are considerable (9.1 and 14.2 absolute points respectively).
A possible reason for this is that embedding based methods map every link in the KG into the same embedding space, which implicitly encodes the connectivity of the whole graph. In contrast, path based models use the discrete representation of a KG as input, and therefore have to leave out a significant proportion of the combinatorial path space by selection. For some path based approaches, computation cost is a bottleneck. In particular, NeuralLP and NTP-$\lambda$ failed to scale to the larger datasets and their results are omitted from the table, as~\citet{minerva} reported. 

Ours is the first multi-hop reasoning approach which is consistently comparable or better than embedding based approaches on all five datasets. The best single model, Ours(ConvE), improves the SOTA performance of path-based models on three datasets (UMLS, Kinship, and FB15k-237) by 4\%, 9\%, and 39\% respectively. On WN18RR and NELL-995, our approach did not significantly improve over existing SOTA. The NELL-995 dataset consists of only 12 relations in the test set and, as we further detail in the analysis (\S~\ref{subsec:rel_type}), our approach is less effective for those relation types.

The model variations using different reward shaping modules perform similarly. While a better reward shaping module typically results in a better overall model, an exception is WN18RR, where ComplEx performs slightly worse on its own but is more helpful for reward shaping. We left the study of the relationship between the accuracy of the reward shaping module and the overall model performance as future work.

\subsection{Ablation Study} 
\label{subsec:ablation}

\begin{table}[t]
    \setlength\tabcolsep{1pt}
    \scalebox{0.8}{
    \begin{tabular}{c|c|c|c|c|c}
    	\toprule
        Model & UMLS & Kinship & FB15k237 & WN18RR & NELL995 \\
        \midrule
        Ours(ConvE) & \highest{73.0} & \highest{75.0} & \highest{38.2} & 43.8 & \highest{78.8} \\
        $-$RS & 67.7 & 66.5 & 35.1 & \highest{45.7} & 78.4 \\
        $-$AD  & 61.3 & 65.4 & 31.0 & 39.1 & 76.1 \\
        \bottomrule
	\end{tabular}
    }
    \caption{Comparison of dev set MRR of Ours(ConvE) and models without reward shaping and action dropout.}
    \label{tb:ablation}
\end{table}

We perform an ablation study where we remove reward shaping ($-$RS) and action dropout ($-$AD) from Ours(ConvE) and compare their MRRs to the whole model on the dev sets.\footnote{According to Table~\ref{tb:ablation} and Table~\ref{tb:mainresults}, the dev and test set evaluation metrics differ significantly on several datasets. We discuss the cause of this in \S~\ref{ap:results_full_kg}.}
As shown in Table~\ref{tb:ablation}, on most datasets, removing each component results in a significant performance drop. The exception is WN18RR, where removing the ConvE reward shaping module improves the performance.\footnote{A possible explanation for this is that as path-based models tend to outperform the embedding based approaches on WN18RR, ConvE may be supplying more noise than useful information about the KG. Yet counter-intuitively, we found that adding the ComplEx reward shaping module helps, despite the fact that ComplEx performs slightly worse than ConvE on this dataset. This indicates that dev set accuracy is not the only factor which determines the effectiveness of reward shaping.} Removing reward shaping on NELL-995 does not change the results significantly.
In general, removing action dropout has a greater impact, suggesting that thorough exploration of the path space is important across datasets.

\subsection{Analysis}
\label{subsec:analysis}

\begin{figure*}[t]
	\centering
	\includegraphics[width=\linewidth]{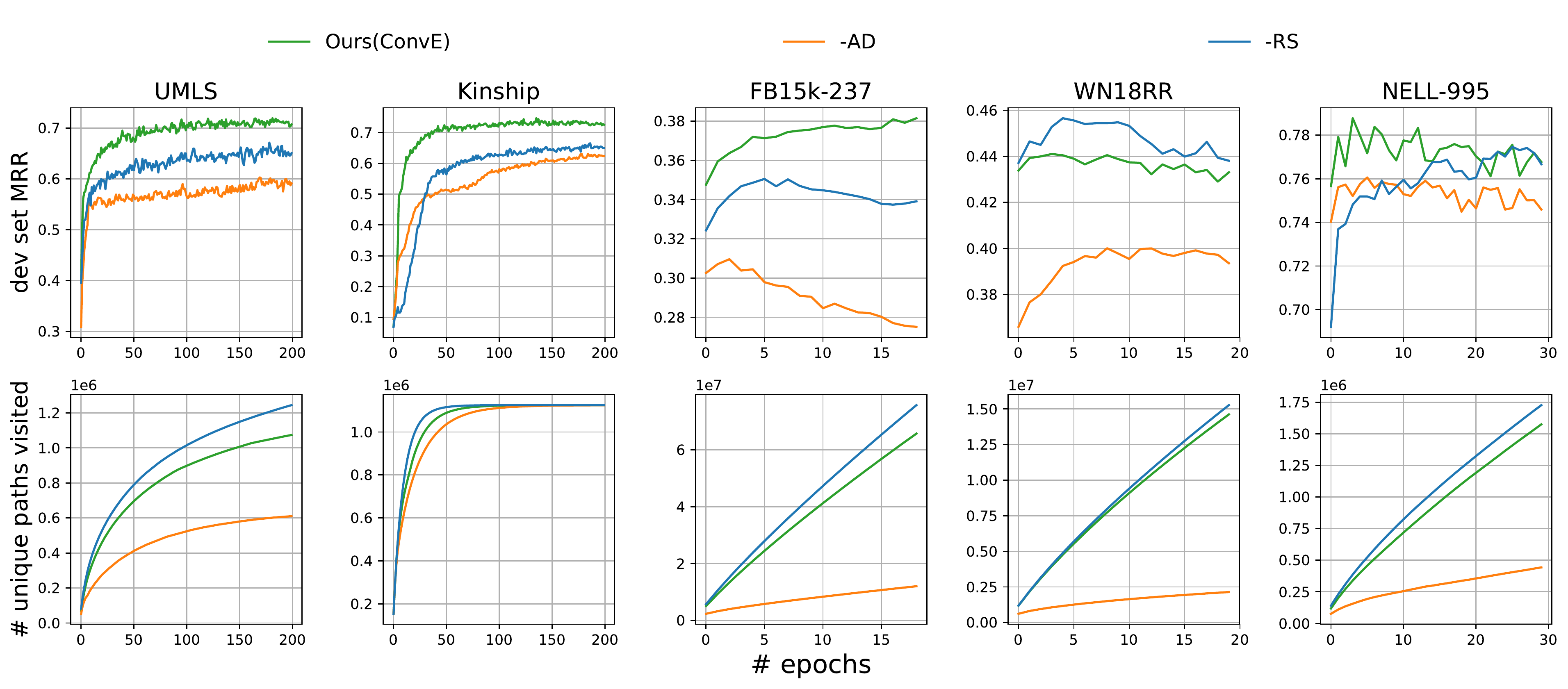}
	\caption{
    	Illustration of convergence rate and path exploration efficiency. The three curves in each subplot represents Ours(ConvE) (green) and the two ablated models: $-$RS (blue) and $-$AD (orange). The top row shows the change of dev set MRR and the bottom row shows the growth of \# unique paths explored w.r.t. \# epochs.
    } 
	\label{fig:analysis}
\end{figure*} 

\subsubsection{Convergence Rate}
We are interested in studying the impact of each proposed enhancement on the training convergence rate. 
In particular, we expect reward shaping to accelerate the convergence of RL (to a better performance level) as it propagates prior knowledge about the underlying KG to the agent. 
On the other hand, a fair concern for action dropout is that it can be slower to train, as the agent is forced to explore a more diverse set of paths. Figure~\ref{fig:analysis} eliminates this concern.

The first row of Figure~\ref{fig:analysis} shows the changes in dev set MRR of Ours(ConvE) (green) and the two ablated models w.r.t. \# epochs. In general, the proposed approach is able to converge to a higher accuracy level much faster than either of the ablated models and the performance gap often persists until the end of training (on UMLS, Kinship, and FB15k-237). Particularly, on FB15k-237, our approach still shows improvement even after the two ablated models start to overfit, with $-$AD beginning to overfit sooner. On WN18RR, introducing reward shaping hurt dev set performance from the beginning, as discussed in \S~\ref{subsec:ablation}. On NELL-995, Ours(ConvE) performs significantly better in the beginning, but $-$RS gradually reaches a comparable performance level. 

It is especially interesting that introducing action dropout immediately improves the model performance on all datasets. A possible explanation for this is that by exploring a more diverse set of paths the agent learns search policies that generalize better.

\begin{table*}[htbp!]
	\centering
    \scalebox{0.85}{
	\begin{tabular}{l|r|rrr|r|rrr}
    	\hline
    	\tworow{\textbf{Dataset}} & \fourcol{\textbf{To-many}} & \fourcolnb{\textbf{To-one}} \\
        \cline{2-5}\cline{6-9}
        & \% & Ours(ConvE) & $-$RS & $-$AD & \% & Ours(ConvE) & $-$RS & $-$AD \\
        \hline
        {UMLS} & 99.1 & 73.1 & \rela{67.9}{\highest{-7}} & \rela{61.3}{\highest{-16}} & 0.9 & 62.5 & \rela{55.5}{\highest{-11}} & \rela{54.4}{\highest{-13}} \\
        {Kinship} & 100 & 75 & \rela{66.5}{\highest{-11}} & \rela{65.4}{\highest{-13}} & 0 &  -- & -- & -- \\
        {FB15k-237} & 76.6 & 28.3 & \rela{24.5}{\highest{-13}} & \rela{20.9}{\highest{-26}} & 23.4 & 72 & \rela{69.8}{\highest{-3}} & \rela{63.9}{\highest{-11}} \\
        {WN18RR} & 52.8 & 65 & \rela{65.7}{+1} & \rela{57.9}{\highest{-11}} & 47.2 & 20.1 & \rela{23.2}{+16} & \rela{18.1}{\highest{-10}} \\
        {NELL-995} & 12.9 & 55.7 & \rela{62.1}{+12} & \rela{56.9}{+2} & 87.1 & 81.4 & \rela{80.7}{\highest{-1}} & \rela{80.5}{\highest{-1}} \\
        \hline
    \end{tabular}
    }
    \caption{MRR evaluation of different relation types (to-many vs. to-one) on five datasets. The \% columns show the percentage of examples of each relation type found in the development split of the corresponding dataset.  In general, our proposed techniques improve the prediction results for to-many relations more significantly.\label{tb:result_by_relation}}
\end{table*}

\subsubsection{Path Diversity}
We also compute the total number of unique paths the agent explores during training and visualize its change w.r.t. \# training epochs in the second row of Figure~\ref{fig:analysis}. When counting a unique path, we include both the edge label and intermediate entity. First we observe that, on all datasets, the agent explores a large number of paths before reaching a good performance level. The speed of path discovery slowly decreases as training progresses. On smaller KGs (UMLS and Kinship), the rate of encountering new paths is significantly lower after a certain number of epochs, and the dev set accuracy plateaus correspondingly. On much larger KGs (FB15k-237, WN18RR, and NELL-995), we did not observe a significant slowdown before severe overfitting occurs and the dev set performance starts to drop. A possible reason for this is that the larger KGs are more sparsely connected compared to the smaller KGs (Table~\ref{tb:data}), therefore it is less efficient to gain generalizable knowledge from the KG by exploring a limited proportion of the path space through sampling.  

Second, it is interesting to see that while removing action dropout significantly lowers the effectiveness of path exploration (orange vs. green), removing reward shaping slightly improves the \# paths visited during training for all datasets. This indicates that the correlation between \# paths explored and dev set performance is not strictly positive. The best performing model in general is not the model that explored the largest \# paths. It also demonstrates the role of reward shaping as a regularizer which guides the agent to avoid noisy paths with its prior knowledge. 

\begin{table*}[htbp!]
    \centering
	\scalebox{0.85}{
	\begin{tabular}{l|r|rrr|r|rrr}
    	\hline
    	\tworow{\textbf{Dataset}} & \fourcol{\textbf{Seen Queries}} & \fourcolnb{\textbf{Unseen Queries}} \\
        \cline{2-5}\cline{6-9}
        & \% & Ours(ConvE) & $-$RS & $-$AD & \% & Ours(ConvE) & $-$RS & $-$AD \\
        \hline
        {UMLS} & 97.2 & 73.1 & \rela{67.9}{\highest{-7}} & \rela{61.4}{\highest{-16}} & 2.8 & 68.5 & \rela{61.5}{\highest{-10}} & \rela{58.7}{\highest{-14}} \\
        {Kinship} & 96.8 & 75.1 & \rela{66.5}{\highest{-11}} & \rela{65.8}{\highest{-12}} & 3.2 & 73.6 & \rela{64.3}{\highest{-13}} & \rela{53.3}{\highest{-27}} \\
        {FB15k-237} & 76.1 & 28.3 & \rela{24.3}{\highest{-14}} & \rela{20.6}{\highest{-27}} & 23.9 & 70.9 & \rela{69.1}{\highest{-2}} & \rela{63.9}{\highest{-10}} \\
        {WN18RR} & 41.8 & 60.8 & \rela{62.0}{+2} & \rela{53.4}{\highest{-12}} & 58.2 & 31.5 & \rela{33.9}{+7} & \rela{28.8}{\highest{-9}} \\
        {NELL-995} & 15.3 & 40.4 & \rela{45.9}{+14} & \rela{42.5}{+5} & 84.7 & 85.5 & \rela{84.7}{\highest{-1}} & \rela{84.3}{\highest{-1}} \\
        \hline
    \end{tabular}
    }
    \caption{MRR evaluation of seen queries vs. unseen queries on five datasets. The \% columns show the percentage of examples of seen/unseen queries found in the development split of the corresponding dataset.\label{tb:result_by_sq}}
\end{table*}
\subsubsection{Performance w.r.t. Relation Types}
\label{subsec:rel_type}

We investigate the behaviors of our proposed approach w.r.t different relation types. For each KG, we classify its set of relations into two categories based on the answer set cardinality. Specifically, we define the metric $\xi_r$ as the average answer set cardinality of all queries with topic relation $r$. We count $r$ as a ``to-many'' relation if $\xi_r>1.5$, which indicates that most queries in relation $r$ has more than 1 correct answer; we count $r$ as a ``to-one'' relation otherwise, meaning most queries of this relation have only 1 correct answer. 

Table~\ref{tb:result_by_relation} shows the percentage of examples of to-many and to-one relations on each dev dataset and the MRR evaluation metrics of previously studied models computed on the examples of each relation type.
Since UMLS and Kinship are densely connected, they almost exclusively contain to-many relations. FB15k-237 mostly contains to-many relations. In Figure~\ref{fig:analysis}, we observe the biggest relative gains from the ablated models on these three datasets. WN18RR is more balanced and consists of slightly more to-many relations than to-one relations. The NELL-995 dev set is a unique one which almost exclusively consists of to-one relations. There is no common performance pattern over the two relation types across datasets: on some datasets all models perform better on to-many relations (UMLS, WN18RR) while others reveal the opposite trend (FB15k-237, NELL-995). We leave the study of these differences to future work. 

We show the relative performance change of the ablated models $-$RS and $-$AD w.r.t. Ours(ConvE) in parentheses. We observe that in general our proposed enhancements are effective in improving query-answering over both relation types (more effective for to-many relations). 
However, adding the ConvE reward shaping module on WN18RR hurts the performance over both to-many and to-one relations (more for to-one relations). 
On NELL-995, both techniques hurt the performance over to-many relations.

\subsubsection{Performance w.r.t. Seen Queries vs. Unseen Queries}

Since most benchmark datasets randomly split the KG triples into train, dev and test sets, the queries that have multiple answers may fall into multiple splits. As a result, some of the test queries $(e_s, r_q, ?)$ are seen in the training set (with a different set of answers) while the others are not. We investigate the behaviors of our proposed approach w.r.t. seen and unseen queries. 

Table~\ref{tb:result_by_sq} shows the percentage of examples associated with seen and unseen queries on each dev dataset and the corresponding MRR evaluation metrics of previously studied models. On most datasets, the ratio of seen vs. unseen queries is similar to that of to-many vs. to-one relations (Table~\ref{tb:result_by_relation}) as a result of random data split, with the exception of WN18RR. On some datasets, all models perform better on seen queries (UMLS, Kinship, WN18RR) while others reveal the opposite trend. 
On NELL-995 both of our proposed enhancements are not effective over the seen queries. 
We leave the study of these model behaviors to future work. In most cases, our proposed enhancements improve the performance over unseen queries, with AD being more effective. 

\section{Conclusions}

We propose two modeling advances for end-to-end RL-based knowledge graph query answering: reward shaping and action dropout. Our approach improves over state-of-the-art multi-hop reasoning models consistently on several benchmark KGs. A detailed analysis indicates that the access to a more accurate environment representation (reward shaping) and a more thorough exploration of the search space (action dropout) are important to the performance boost. 

On the other hand, the performance gap between RL-based approaches and the embedding-based approaches for KGQA remains. In future work, we would like to investigate learnable reward shaping and action dropout schemes and apply model-based RL to this domain. 
\section*{Acknowledgements}
We thank Mark O. Riedl, Yingbo Zhou, James Bradbury and Vena Jia Li for their feedback on early draft of the paper, and Mark O. Riedl for helpful conversations on reward shaping. We thank the anonymous reviewers and the Salesforce research team members for their thoughtful comments and discussions. We thank Fr\'ederic Godin for pointing out an error in Equation~\ref{eq:ad} in an early version of the paper.

\bibliography{emnlp2018}
\bibliographystyle{acl_natbib_nourl}

\clearpage
\newpage

\begin{appendices}

\section{Appendix}
\subsection{Model Robustness}
\label{ap:model_robust}
We run the best embedding-based model ConvE and Ours(ConvE) on all datasets using 5 different random seeds with all other hyperparameters fixed. Table~\ref{tb:model_robustness} reports the mean and standard deviation of each model. We observe that both models demonstrate a small standard deviation ($<0.01$) on all datasets.
\begin{table}[ht!]
    \centering
	\scalebox{1.0}{
	\begin{tabular}{l|c|c}
    	\toprule
        Dataset & ConvE & Ours(ConvE) \\
        \midrule
        UMLS & 95.5$\pm$0.4 & 93.7$\pm$0.2 \\
        Kinship & 86.9$\pm$0.3 & 86.2$\pm$0.7 \\
        FB15k-237 & 43.5$\pm$0.1 & 40.7$\pm$0.2 \\
        WN18RR & 45.3$\pm$0.4 & 44.7$\pm$0.2 \\
        NELL-995 & 76.2$\pm$0.3 & 72.7$\pm$0.4 \\
        \bottomrule
    \end{tabular}
    }
    \caption{Test set MRR$\times$100 mean and standard deviation across five runs on all datasets.\label{tb:model_robustness}}
\end{table}

\subsection{Development Set Evaluation Using Complete KGs}
\label{ap:results_full_kg}
Comparing Table~\ref{tb:mainresults} and Table~\ref{tb:ablation} reveals that the dev set MRRs are significantly lower than the test set MRRs on some datasets (UMLS, Kinship and FB15k-237). Such discrepancies are caused by the multi-answer queries in these datasets. As most benchmark datasets randomly split the KG triples into train/dev/test sets, the queries that have multiple answers may fall into multiple splits. Because we hide all triples in the test set during the dev set evaluation, some predictions generated during dev set evaluation were wrongly punished as false negatives. In contrast, the test set evaluation metrics are computed using the complete KGs. Access to the complete KG eliminates most of the false negatives cases and hence increases the performance. 
\begin{table}[htbp!]
    \setlength\tabcolsep{1pt}
    \scalebox{0.8}{
    \begin{tabular}{c|c|c|c|c|c}
    	\toprule
        Model & UMLS & Kinship & FB15k237 & WN18RR & NELL995 \\
        \midrule
        Ours(ConvE) & \highest{95.1} & \highest{86.8} & \highest{41.8} & 44.1 & \highest{78.8} \\
        $-$RS & 85.6 & 75.7 & 37.1 & \highest{46.1} & 78.4 \\
        $-$AD  & 76.2 & 75.9 & 32.4 & 39.3 & 76.1 \\
        \bottomrule
	\end{tabular}
    }
    \caption{Comparison of dev set MRR computed using the complete KGs of Ours(ConvE) and models without reward shaping and action dropout.}
    \label{tb:ablation_fg}
\end{table}

Table~\ref{tb:ablation_fg} shows the dev set MRR of the same systems shown in Table~\ref{tb:ablation} with the MRRs computed using the complete KGs. On four of the datasets, the evaluation metrics increases significantly to the level that is comparable to those on the test set, with the relative improvement correlated with the average node fan-out in the KG (Table~\ref{tb:data}).

Notice that Table~\ref{tb:ablation_fg} is generated after all hyperparameters were fixed and the purpose is to show the effects of such dataset peculiarities. To avoid potential test set leakage, hyperparameter search should be done with the test set triples hidden (Table~\ref{tb:data}) instead of with the full KG.



\subsection{Action Dropout Rates Used for Different KGs}
\label{ap:ad_rates}
Table~\ref{tb:ad_rates} show the action dropout rates used for all KG datasets in our experiments. In general, larger action dropout rates are necessary for KGs that are densely connected. We find a positive correlation between the optimal action dropout rate and the average node fan-out (Table~\ref{tb:data}).

For UMLS and Kinship, we tried setting the action dropout rate to 1.0 (completely random sampling) and observed small but significant performance drop. Random sampling performs reasonably well on these two datasets possibly due to the fact that they are small. For larger KGs (FB15k-237, WN18RR, NELL-995), policy-guided sampling is necessary. 
\begin{table}[ht!]
    \centering
	\scalebox{1.0}{
	\begin{tabular}{l|c}
    	\toprule
        Dataset & $\alpha$ \\
        \midrule
        UMLS & 0.95 \\
        Kinship & 0.9 \\
        FB15k-237 & 0.5 \\
        WN18RR & 0.1 \\
        NELL-995 & 0.1 \\
        \bottomrule
    \end{tabular}
    }
    \caption{Action dropout rates used in our experiments.\label{tb:ad_rates}}
\end{table}

\end{appendices}

\end{document}